# Image anomaly detection with capsule networks and imbalanced datasets[s]

Claudio Piciarelli[0000−0001−5305−1520], Pankaj Mishra[0000−0003−1945−5737], and Gian Luca Foresti[0000−0002−8425−6892]

University of Udine, via delle Scienze 206, 33100 Udine Italy
claudio.piciarelli@uniud.it, mishra.pankaj@spes.uniud.it,
gianluca.foresti@uniud.it

**Abstract.** Image anomaly detection consists in finding images with anomalous, unusual patterns with respect to a set of normal data. Anomaly detection can be applied to several fields and has numerous practical applications, e.g. in industrial inspection, medical imaging, security enforcement, etc.. However, anomaly detection techniques often still rely on traditional approaches such as one-class Support Vector Machines, while the topic has not been fully developed yet in the context of modern deep learning approaches. In this paper we propose an image anomaly detection system based on capsule networks under the assumption that anomalous data are available for training but their amount is scarce.

## 1 Introduction

Anomaly detection has always been a challenging problem in the field of machine learning. It consists in identifying anomalies within datasets, where an anomaly is anything that significantly differs from the majority of the data. Anomaly detection is thus achieved by building a model of "normality" and then comparing any subsequent data with that model.

The topic has many potential application fields, such as identification of defective product parts in industrial vision applications [12], fault-prevention in industrial sensing systems [8], detection of anomalous network activity in intrusion detection systems [1], medical image analysis for tumor detection [3], traffic analysis [16], structural integrity check in hazardous or inaccessible environments [15] and many more.

Many classical machine learning techniques have been adopted to identify anomalies in data [7], such as Bayesian networks, rule-based systems, clustering algorithms, statistical analysis, etc.. One of the most popular approaches relies on Support Vector Machines and in particular on their one-class variant, in which the standard SVM technique is used to split the feature space in two parts, one with high-density data (the normal class) and the other with outliers. Despite this huge interest of the research community on anomaly detection, the topic has not been fully developed in the context of modern deep learning. In this field (which we will call from now on *deep anomaly detection*), relatively few works have been published, mostly relying on reconstruction-based or generative-based approaches. The aim of this paper is to investigate the use of deep learning techniques for image anomaly detection: the task is to search for those images that are visually different from a reference group. In particular, we will focus on a recent evolution of deep learning techniques, the so-called capsule networks [18], to check if they could fit image anomaly detection tasks.

[s] This work is partially supported by beanTech s.r.l. and by the Italian Ministry of Defense project 171/2016 RA²M.

Anomaly detection techniques can be roughly classified in three main groups depending on the availability of data and labels: fully-supervised, semi-supervised and unsupervised [7]. In the *fully-supervised* case, it is assumed that both normal and anomalous data are available for training, and the problem reduces to a standard classification task. In this case, the main difference between anomaly detection and other classification problems is the imbalanced nature of the dataset: anomalies may be available for training, but their amount is by definition much smaller than normal data. In the *semi-supervised* case only normal data is labeled and available for training, and the goal is to classify new data as either normal or anomalous — this is why this approach is often called "novelty detection". Finally, the *unsupervised* case (also called "outlier detection") is similar to a clustering problem: no labels are given for the training set, which could potentially contain both normal and anomalous data, and the goal is to identify the normal cluster while leaving out the outliers. In this paper we will focus on the fully-supervised approach, and a capsule network will be used as regular classifier on imbalanced data.

The paper is organized as follows: in section 2 we give an overview of the most recent works in the field of deep anomaly detection. Section 3 describes our capsule-based architecture and how it has been adapted to the task of anomaly detection. Finally, in section 4 we provide experimental results on several datasets to show the effectiveness of the proposed method.

## 2   Related works

As mentioned in section 1, anomaly detection has been widely studied in the field of classical machine learning. Chandola et al. [7] give an excellent survey on the topic, highlighting the different types of anomalies, application fields, and possible non-deep approaches. From a deep learning point of view, the topic has been less extensively covered. Kiran et al. published a survey on this topic, but it is exclusively focused on anomaly detection in videos [13].

The fully-supervised approach is generally addressed with generic techniques for handling imbalanced data, such as undersampling the dominant class or oversampling the smaller class either by data duplication or by synthetic generation of new data. In both cases the idea is to use a pre-processing step to make the dataset balanced before applying any classification algorithm [4].

Regarding semi-supervised or unsupervised approaches, early works adopted techniques such as Deep Belief Networks [20] for medical diagnosis on EEG waveforms or Restricted Boltzmann Machines [11] for network traffic analysis. More recently, hybrid approaches have been proposed in which deep architectures are used together with ideas from classical machine learning: for example Ruff et al. [17] propose the Deep Support Vector Data Description method, in which a deep neural network is trained under the same constraints adopted by one-class Support Vector Machines. However, the majority of the proposed works currently rely either on deep autoencoders or generative models.

Autoencoders are neural networks in which the differences between the output and the input are minimized: the ideal autoencoder thus is an identity function. However, autoencoders are implemented as a concatenation of an encoder and a decoder with an intermediate bottleneck, a low-dimension layer in which the original data are compressed. If the decoder part can reconstruct the original input, then the latent representation in the bottleneck captures all the relevant features of the original data. Despite autoencoders have been initially developed for dimensionality reduction tasks, they can be adapted to anomaly detection problems: if an autoencoder is trained on the normal class, it will learn how to represent its main features in its latent space. When an anomalous input is fed in the network, it is assumed it cannot be properly represented in the latent space, and thus the decoder reconstruction will be poor [6, 5, 22].

The other main approach is based on generative models, and in particular on generative adversarial networks (GAN). GAN are based on two competing networks: a generator, trying to create new data similar to the training ones, and a discriminator, trying to discern original data from the generated ones. The competition between the two networks leads the generator to learn how to create novel data which are similar to the training ones. This way, if trained on normal data, the generator learns a "normality model" much like autoencoders do. If the generator is inverted, a comparison on the latent representations of normal and anomalous data can be used to detect anomalies [10, 19, 2].

## 3  System description

In this work we address the anomaly detection problem as a fully-supervised classification with highly imbalanced datasets. The model we adopted is a capsule network, in particular we rely on the CapsNet architecture proposed by Hinton in [18].

The original network has been developed to recognize MNIST digits and consists of two main parts: an encoder, converting an input image into 10 vectors of instantiation parameters (digit caps), and a decoder, which reconstructs the original input. The network is trained in order to maximize the vector length of the correct digit caps and to minimize the decoder reconstruction loss. Although the decoder is not strictly necessary, it is used to force the digit caps to learn meaningful instantiation parameters describing visual properties of the digits. The main components of the network are:

- **Convolution:** It is a traditional convolutional layer. The aim is to extract basic features from the image. The network uses 256 kernels of size 9x9 with ReLU output.
- **PrimaryCaps:** This layer is similar to the convolution layer, and it outputs 1152 feature vectors in $R^8$. These vectors are fed to a squash function, which preserves their orientation and normalizes the length in the range $[0, 1]$.
- **Routing by Agreement:** Routing by agreement is somewhat similar to max pooling. It decides what information to send to the next level. In this method each capsule tries to predict the next layer's activations based on its length and orientation.
- **DigitCaps:** After routing by agreement, 10 digit caps are obtained. These squashed vectors in $R^{16}$ represent the instantiation parameters of each digit class. The vector length is proportional to the probability of the input belonging to a specific class, while its orientation represents the "pose", this is the specific instance of a digit among the many possible appearances for the same digit.
- **Reconstruction :** The reconstruction part takes the longest digit caps vector and uses three fully connected layers to reconstruct the input image.

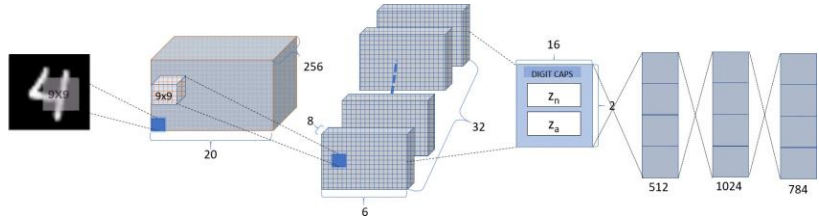

**Fig. 1.** The CapsNet architecture adopted in this work.

In our implementation, we adopt the CapsNet architecture to perform fully-supervised anomaly detection, and thus we reduced the number of digit caps to 2: one representing the normal class, and the other representing the anomaly class, as shown in Figure 1. However, in section 4 we will show that this basic network has extremely poor performances when the dataset is highly imbalanced. In order to deal with the class imbalance, we adopted two anomaly measures: reconstruction loss and vector length difference.

*Reconstruction loss $r_l$* is a MSE loss computed on the difference between original and reconstructed image. We force the decoder network to be trained only on normal data and using the output of the normal digit caps. This way, the network will be able to reconstruct correctly only normal data, and it will behave poorly on anomalous data. This is the same technique adopted by nearly all the autoencoder-based methods described in section 2.

*vector length difference* uses the length of digit caps vectors as a measure of anomaly. Let $z_n$ and $z_a$ the two output vectors for normal and anomaly classes. Using the standard CapsNet approach, an image is classified as anomaly if $\|z_a\| > \|z_n\|$, but this approach does not give good results on imbalanced datasets. With imbalanced datasets we noticed that the system behaves as expected on the dominant class ($\|z_n\| \approx 1, \|z_a\| \approx 0$), while on anomalous data the difference between the two vectors lengths is typically smaller. For example, $\|z_n\| = 0.8, \|z_a\| = 0.6$ is a strong hint that the sample is anomalous, even though it would be classified as normal from a standard CapsNet. We thus propose to use $\|z_a\| - \|z_n\|$ as anomaly score.

The final anomaly score $AS$ is a combination of the two measures:

$$AS = \|z_a\| - \|z_n\| + r_l \tag{1}$$

with $\|z_a\|, \|z_n\|, r_l \in [0, 1]$. The ROC curve in Figure 2 shows that the combination of the two anomaly measures leads to better results than using ony one of the two. Once computed the anomaly score on the training data, it is fed into a logistic regressor to find the optimal threshold separating normal from anomalous data. The threshold can later be used to classify new data based on their anomaly score (see Figure 3).

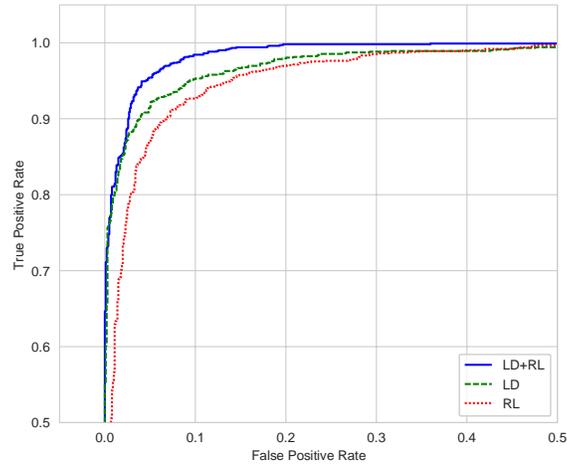

**Fig. 2.** ROC curve for three anomaly detection measures: vector length difference, reconstruction loss, and vector length difference + reconstruction loss.

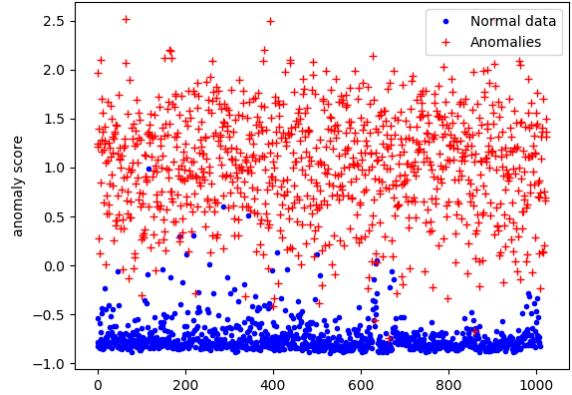

**Fig. 3.** Anomaly scores on test data, training done with 10% anomalies. Logistic regression threshold: -0.09.

## 4   Results

The proposed method has been evaluated on three datasets (MNIST [14], Fashion-MNIST [21], Kuzushiji-MNIST [9]). Each dataset has 10 classes, respectively representing digits, clothing and ancient Japanese characters. Training has been performed by iterating the following schema over all classes:

1. Choose a class as the normal class
2. The training dataset contains all the training images of the chosen class plus some training images randomly picked from the other classes. The amount of training anomalies is either 10% or 1%
3. Train the network and compute the anomaly score threshold
4. Test the system on the whole test dataset

Note that the test dataset is not imbalanced, this avoids biased accuracy results. Table 1 shows the training hyperparameters.

**Table 1.** Training hyperparameters

| Adam learning Rate | 0.001 |
|---|---|
| % of anomalies in training data | 1% – 10% |
| batch size | 32 |
| Epochs | 10 |

**MNIST Dataset**: For the MNIST dataset the network is trained on $28 \times 28 \times 1$ MNIST digit images. The images have been standardized with mean 0.1307 and std. deviation 0.3081. Table 2 shows the achieved results with a standard CapsNet and with the proposed approach, in the cases of 10% and 1% of anomalies in the training set. As it can be seen, the standard CapsNet approach fails when the dataset is extremely imbalanced: when anomalies are 1% of the training dataset, the standard CapsNet has an average accuracy of 51.44%, which is very close to a random guess. On the other hand, the proposed system keeps a high accuracy even with imbalanced training data (accuracy is on average 98.84% and 96.46% for the 10% and 1% anomaly cases respectively). Figure 4 shows the reconstructed images for both normal and anomalous data. The figure confirms that reconstruction is poor on anomalies, thus motivating the use of reconstruction error in the anomaly score definition.

**Table 2.** Accuracy % on MNIST dataset for standard CapsNet and the proposed method. The amount of anomalies in the training data is 10% (top rows) or 1% (bottom rows).

|  | 0 | 1 | 2 | 3 | 4 | 5 | 6 | 7 | 8 | 9 | avg |
|---|---|---|---|---|---|---|---|---|---|---|---|
| Standard, 10% an. | 97.46 | 98.78 | 97.02 | 92.87 | 96.36 | 93.42 | 96.87 | 96.83 | 95.50 | 92.13 | 95.72 |
| Proposed, 10% an. | **99.50** | **99.27** | **99.22** | **99.21** | **99.10** | **98.33** | **98.74** | **98.05** | **99.00** | **97.93** | **98.84** |
| Standard, 1% an. | 48.90 | 73.58 | 50.00 | 49.66 | 48.95 | 46.56 | 48.34 | 50.00 | 48.75 | 49.63 | 51.44 |
| Proposed, 1% an. | **99.20** | **98.24** | **98.19** | **95.48** | **94.37** | **95.46** | **98.34** | **97.07** | **97.85** | **90.41** | **96.46** |

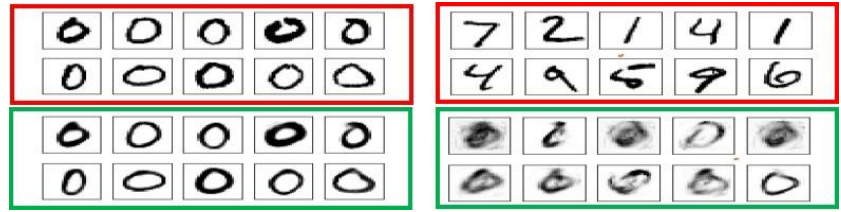

**Fig. 4.** Top rows: normal (left) and anomalous (right) samples from the MNIST test set. Bottom rows: the reconstructed images.

**Fashion MNIST Dataset**: Fashion MNIST dataset is composed of images from an online clothing store. it contains 60,000 examples as a training set and 10,000 examples as a test set organized in 10 classes (see Table 3). The images are 28× 28 grayscale images and have been standardized as in the MNIST case. Results are shown in Table 4 and reconstructions are shown in Figure 5. The dataset is more challenging, but the results confirm that the proposed method outperforms standard CapsNet when the number of training anomalies is small.

**Kuzushiji-MNIST(K-MNIST)**: It is a dataset of 28 ×28 grayscale images of ancient Japanese handwritten characters. The dataset contains 60,000 images for training and 10,000 images for testing. Images have been standardized before processing. It is a challenging dataset, as it can be seen in Figure 6, where the 10 rows corresponding to each class can be seen. The accuracy for K-MNIST dataset can be seen in table 5 and reconstruction examples are in Figure 7. The results obtained on other datasets are confirmed: the proposed method outperforms standard capsule network classification, especially in the 1% training anomaly case.

**Table 3.** Fashion MNIST label encoding

| Label | 0 | 1 | 2 | 3 | 4 | 5 | 6 | 7 | 8 | 9 |
|---|---|---|---|---|---|---|---|---|---|---|
| Desc. | T-shirt/top | Trouser | Pullover | Dress | Coat | Sandal | Shirt | Sneaker | Bag | Ankle boot |

**Table 4.** Accuracy % on Fashion MNIST dataset.

|  | 0 | 1 | 2 | 3 | 4 | 5 | 6 | 7 | 8 | 9 | avg |
|---|---|---|---|---|---|---|---|---|---|---|---|
| Standard, 10% an. | 88.14 | 96.89 | 85.08 | 92.19 | 85.77 | 96.15 | 76.93 | 95.11 | 94.96 | 97.18 | 90.84 |
| Proposed, 10% an. | **93.28** | **98.07** | **87.50** | **95.01** | **91.50** | **98.07** | **84.44** | **96.64** | **97.73** | **97.83** | **94.01** |
| Standard, 1% an. | 49.41 | 49.41 | 49.41 | 49.41 | 49.41 | 49.41 | 49.41 | 49.46 | 49.41 | 49.41 | 49.41 |
| Proposed, 1% an. | **87.45** | **95.31** | **84.98** | **90.86** | **87.70** | **94.27** | **77.32** | **93.33** | **92.14** | **96.15** | **89.95** |

## 5  Conclusions

In this work we proposed a fully-supervised deep anomaly detection technique based on capsule networks. The network is trained as in a binary classification problem, where each sample is either normal or anomalous, but with the additional constraint of imbalanced datasets. To deal with data imbalance, we proposed a novel anomaly score based on output vectors length difference and reconstruction error. Experimental results are very promising, since the network has state-of-the-art performance even with highly imbalanced datasets where the standard network fails.

To the best of our knowledge, this is the first use of capsule networks for anomaly detection tasks. We believe that the ability of capsule networks to create equivariant models can boost anomaly detection in the same way it has proven to boost standard classification problems.

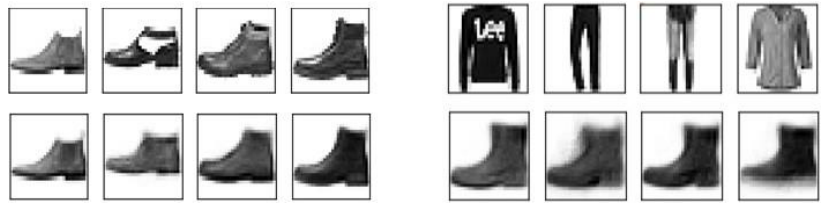

**Fig. 5.** Top row: normal (left) and anomalous (right) samples from the Fashion MNIST test set. Bottom row: the reconstructed images.

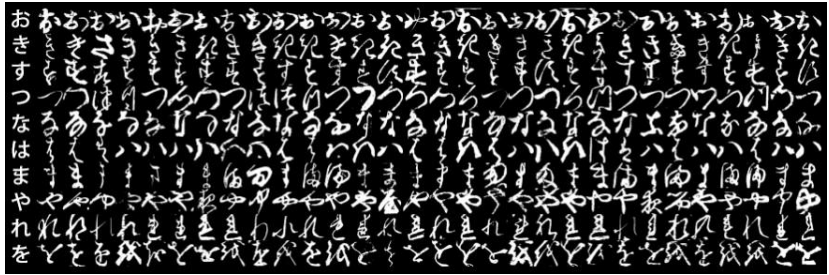

**Fig. 6.** 10 classes of Kuzushiji-MNIST, with the first column showing each character's modern hiragana counterpart.

**Table 5.** Accuracy % on K-MNIST dataset.

|  | 0 | 1 | 2 | 3 | 4 | 5 | 6 | 7 | 8 | 9 | avg |
|---|---|---|---|---|---|---|---|---|---|---|---|
| Standard, 10% an. | 91.55 | 80.58 | 72.88 | 87.40 | 49.41 | 87.10 | 83.05 | 93.23 | 81.92 | 81.97 | 80.91 |
| Proposed, 10% an. | **96.54** | **93.92** | **88.69** | **96.25** | **88.19** | **93.43** | **93.08** | **93.48** | **95.85** | **95.01** | **93.44** |
| Standard, 1% an. | 49.95 | 49.95 | 49.95 | 49.95 | 49.95 | 49.95 | 49.95 | 49.95 | 49.95 | 49.95 | 49.95 |
| Proposed, 1% an. | **92.31** | **86.11** | **79.17** | **93.06** | **83.27** | **90.81** | **86.56** | **76.37** | **87.91** | **90.71** | **86.63** |

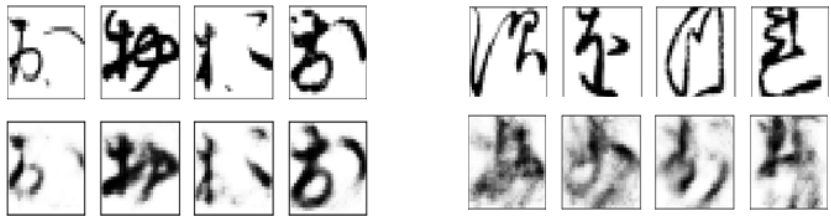

**Fig. 7.** Top row: normal (left) and anomalous (right) samples from the K-MNIST test set. Bottom row: the reconstructed images.

The proposed method currently outperforms or it is comparable to other deep learning anomaly detection techniques as the ones discussed in section 2, however a direct comparison would be unfair since most of those methods use semi-supervised or unsupervised techniques. Fully-supervised anomaly detection is a very relevant topic with many practical applications in which anomalous data are available, but of course semi-supervised or unsupervised approaches are more challenging and can deal with those problems where anomalous data are not available or not labeled. For this reason, as a future work we plan to investigate the use of capsule networks in this direction.